%% file: neurips_2025.tex
\documentclass{article}

\usepackage[dblblindworkshop,final]{neurips_2025}

\usepackage[utf8]{inputenc} 
\usepackage[T1]{fontenc}    
\usepackage{hyperref}       
\usepackage{url}            
\usepackage{booktabs}       
\usepackage{amsfonts}       
\usepackage{nicefrac}       
\usepackage{microtype}      
\usepackage{xcolor}         
\usepackage{algorithm}
\usepackage{algpseudocode}
\usepackage{amsmath}
\usepackage{graphicx} 
\usepackage{pifont}
\usepackage{enumitem}
\newcommand{\cmark}{\ding{51}} 
\newcommand{\xmark}{\ding{55}} 
\newcommand{\warnmark}{\ding{115}} 
\title{COLA: Continual Learning via Autoencoder Retrieval of Adapters}

\author{%
  Jaya Krishna Mandivarapu \\
  Microsoft\\
  \texttt{jmandivarapu1@microsoft.com} \\
}

\begin{document}

\maketitle

\input{Sections/0_abstract}

\input{Sections/1_intro}

\input{Sections/2_Methodology}
\input{Sections/3_ExpANDresults}

\input{Sections/2b_Related_work}
\input{Sections/5_conclusion}
\bibliography{main}
\end{document}

%% file: Sections/0_abstract.tex
\begin{abstract}
Learning a set of tasks over time, also known as continual learning (CL), is one of the
most challenging problems in artificial intelligence due to catastrophic forgetting. Large language models (LLMs) are often impractical to frequent re-training and continual learning , due to high cost of computational resources for training. Moreover, LLM are not suitable for continual learning as updating these models over time for acquiring new knowledge leads to overwrites existing knowledge leading to common phenomenon know as \textit{catastrophic forgetting}.  In this paper, we aim to address these concerns using a novel framework , COLA that employs an autoencoder to learn capture low-dimensional embeddings of the weights associated with various tasks. Our approach facilitates the transfer of knowledge to new tasks while preventing catastrophic forgetting, all without using data replay or a substantial set of task-specific parameters. Our approach, COLA, makes the LLM efficiently learn new tasks with minimal training, insignificant performance degradation on previous tasks, and eliminates the need for retaining earlier training data. Empirical evaluation on different datasets ranging from task oriented dialouge system to intent classsfication datasets showcases that our method not only overcomes catastrophic forgetting but also achieves significant reduction in  parameter usage and memory size, across multiple tasks and outperforming the existing state of the art methods across multiple datasets.



\end{abstract}

%% file: Sections/1_intro.tex
\section{Introduction}
\label{sec:intro}
Lifelong or continual learning (CL) is one of the most challenging problems in field of AI, posing a  significant hurdle in the quest for artificial general intelligence (AGI) \cite{goodfellow2013empirical,kemker2018measuring}.  In this paradigm, a single system should continually learn to solve new tasks without forgetting previously learned tasks, while also effectively using the acquired knowledge from previous knowledge to solve new tasks. In recent times. Two primary challenges in continual learning are: (1) $\textit{preventing catastrophic forgetting}$—where previously acquired knowledge is lost upon learning new tasks \cite{mccloskey1989catastrophic,ratcliff1990connectionist} and (2) $\textit{enabling forward transfer}$, which uses earlier task knowledge to speed up and improve learning on subsequent tasks.  (3) $\textit{promoting backward transfer}$, in which mastering new tasks enhances performance on previously learned tasks

Large Language Models have demonstrated incredible capabilities in absorbing and storing vast amounts of data within their parameters when trained on word knowledge, thus acquiring knowledge from the data. These Pre-trained LLM's has shown promising results by utilizing the knowledge encoded in them. Additionally, these models also exhibited promising results when finetuned on Knowledge Intensive Language Tasks(KILT) eg: Open Domain Question Answering, Slot Filling, Dialogue systems..etc.

One prominent application of KILT-tuned LLMs is in Task Oriented dialogue system which power modern LLM based assistants like  Microsoft Coiplot, Siri and Alexa. These systems are trained to achieve specific objectives in various fields ranging from customer support agents to autonomous AI agents. Traditinally Task-oriented dialogue framework systems follows a modular pipeline: Natural Language Understanding (NLU) for understanding the user's intent, Dialogue State Tracking (DST) for tracking the current dialogue states, Dialogue Policy (DP) and Natural Language Generation (NLG) for generating the responses based on previous conversations. When these modules are trained and optimized individually, can yield high performance on individual subtasks. More researchers have begun to train task oriented dialogue systems end to end with pre-trained Large Language Models(LLM).  However, these monolithic models not only demand extensive annotated data but also suffer from parameter-sharing constraints: updating the entire model to accommodate new scenarios often causes catastrophic forgetting of previously learned tasks. . Due to these limitations, extending end-to-end model to new domains and functionalities often result in high computational cost and operational costs. Therefore , the capability of the model to learn new tasks continuously without forgetting the older tasks a.k.a Continual Learning (CL) is critical for developing a efficient task oriented dialogue systems.

\begin{figure}[t]
  \centering
   \includegraphics[width=1.0\linewidth]{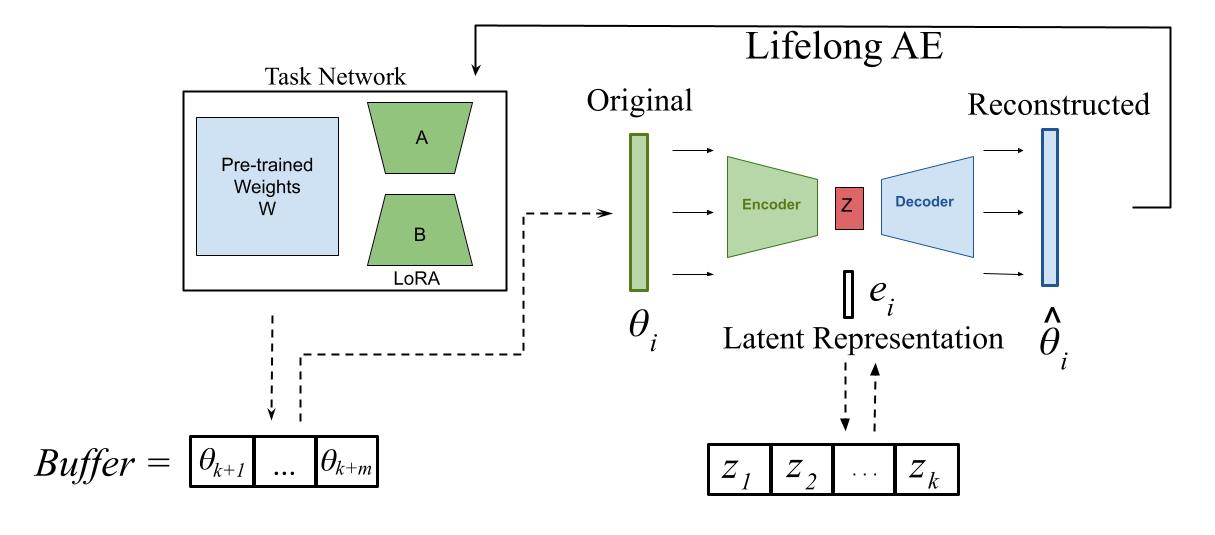}
    \caption{COLA: Our proposed framework consists of two primary components: task-specific adapter networks (TN), an optional buffer for temporarily storing the most recent $m$ tasks, and a lifelong autoencoder (AE) designed for long-term learning. Given a new task $t_k$. we first train a model with parameters $\theta_k$, We learn the optimal weights for each task and temporarily store the trained adapter in a buffer; once full, the stored adapters are encoded into the autoencoder and the buffer is cleared. Alternatively, adapters can be encoded directly without buffering. When an earlier task reoccurs (e.g., revisiting a previously learned task), we reconstruct its adapter weights by decoding the corresponding latent vector \( z_i \) from the autoencoder and load them onto the task-specific adapter. Although the latent vector \( z_i \) is asymptotically smaller than the original parameter set \( \theta_i \), the reconstructed adapter achieves performance closely matching that of the original model.
    }
   \label{fig:framework}
\end{figure}


%% file: Sections/2_Methodology.tex
\section{Methodology}
\label{sec:methodology}

\subsection{Problem Statement}
Continual learning (CL) covers a range of scenarios characterized by specific learning constraints, such as limited memory, fixed model architecture, and restricted access to previously encountered data. In this work, we consider a setting that aligns closely with realistic sustems such as classification, task-oriented dialogue systems ..etc, defined by the following constraints:
\begin{enumerate}[label=\arabic*)]
\item \textbf{Sequential acquisition.}  System learns tasks in a sequential manner. Eg: New Taks (e.g., dialogue tasks, intention classification) arrive one after another; at any moment, the learner observes data from the \emph{only one} task.
\item \textbf{Task separability.} All the task can be learned without relying on information from other tasks (e.g. domain‑specific intent recognition).
\item \textbf{Task awareness.}  During training the system is told which task is active, so task inference is not part of the challenge during training. During evaluation task labels are not available and task inference must be done by the system itself.
\item \textbf{No rehearsal budget.}  Once training on task~$i$ is complete, System can't access task~$i$ dataset or any of the previous tasks. System inherently needs to learn new task without forgetting old tasks prevent catastrophic forgetting without replaying old data..
\end{enumerate}

The above differs fundamentally from class‑incremental(CIL) setting, where a \emph{single} classification head must accommodate an ever‑growing label set.  Instead, our learner faces a sequence of \emph{domain‑isolated} problems and must preserve knowledge of each without revisiting past data.

Formally, each task task~$T_i$  in the CL scenarios is represented by the dataset $\mathcal{D}_i = {(\mathbf{x}^{(j)}, y^{(j)})}{j=1}^{n_i}$, where $\mathbf{x}$ is a input and $y$ is its output label.  The CL system is sequentially exposed to datasets  $(\mathcal{D}_1, \mathcal{D}_2, \dots , \mathcal{D}_k)$ seen exactly once during the system's lifetime.  Multiple passes over $\mathcal{D}_i$ are permitted while it is resident in memory, but no element of $\mathcal{D}_i$ may be retained when the learner moves on to $\mathcal{D}_{i+1}$.

In addressing this constrained scenario, prior approaches typically fall into two categories:  \emph{Regularization} approaches aim for a \textit{single} shared parameter set that is nudged to remain useful for old tasks, and \emph{Expansion based} methods, which allocate a fresh parameter to each new task either by freeing old parameters or adding completely new set of parameters. The former methods offer storage efficiency at the cost of significant performance degradation over time, whereas the latter approaches achieve higher performance but suffer from linearly growing parameter storage requirements.

\textbf{Contributions.}  
We introduce a novel architectural continual learning method that strikes a balance between full-model retraining and expensive rehearsal techniques. Our contributions are threefold:  
\begin{enumerate}[label=(\roman*)]
    \item We augment a fixed pretrained backbone with lightweight, task-specific adapter modules, effectively isolating new task knowledge while preserving previously acquired capabilities—without relying on raw-data rehearsal.  
    \item Upon completion of training a task, we propose to compress adapter weights using a Contractive Autoencoder (CAE), retaining only a compact latent representation and discarding the original adapter parameters. This significantly reduces storage overhead, achieving orders-of-magnitude memory savings.  
    \item The compressed latent codes enable efficient, on-demand decoding to reconstruct task-specific adapters with minimal performance loss, thereby satisfying strict no-rehearsal constraints and overcoming the prohibitive storage costs of replay buffers or full model copies.
    \item Our latent encoding enables transfer learning by selecting and reusing the adapter with the lowest perplexity on new inputs, speeding up training and transferring knowledge across related tasks.

\end{enumerate}

\subsection{Problem Formulation}

\subsection{LoRA Adapters for Parameter-Efficient Continual Learning}
\label{sec:lora}

Large language models (LLMs) now contain billions of parameters, making full fine-tuning prohibitive in both memory and compute. Fine-tuning a large language model(LLM) which contains billions of parameters as s backbone for every new task is unfeasible for continual learning scenarios. Instead, we employ \textbf{LoRA} (low-rank adaptation) adapter as our task-specific module of choice.
LoRA posits that the \emph{update} required to steer a pre-trained weight matrix toward a new task occupies a low-dimensional subspace.  \textbf{LoRA}—Low-Rank Adaptation—addresses this by extending the network by injecting \emph{small, trainable rank-decomposition matrices} into frozen projection layers of the Transformer \cite{hu2021loralowrankadaptationlarge}. When compared to traditional complete model fine-tuning, LoRA reduces the overall trainable parameters by up to four orders of magnitude and cuts GPU memory by a factor of three while achieving the similar task accuracy on GPT-2 \cite{radford2019language}, RoBERTa \cite{liu2019roberta}, and GPT-3 \cite{brown2020language} benchmarks.

\vspace{0.4em}
\noindent\textbf{Low-rank update:} Let's consider  $\mathbf{W}_{0},\mathbf{W}_{1},\mathbf{W}_{2} ... \mathbf{W}_{n}$ be the weights of each layer of n-layer LLM model. For a given layer $\mathbf{W}_{0}\!\in\!\mathbb{R}^{d\times k}$ be a frozen projection
matrix (e.g.\ the query or value matrix inside self-attention).
Rather than learning a full update
$\Delta\mathbf{W}\!\in\!\mathbb{R}^{d\times k}$,
LoRA constrains it to a rank-$r$ factorisation:
\begin{equation}
    \mathbf{W}
    \;=\;
    \mathbf{W}_{0} \;+\; \Delta\mathbf{W}
    \quad\text{with}\quad
    \Delta\mathbf{W}
    \;=\;
    \mathbf{B}\,\mathbf{A},
    \label{eq:lora}
\end{equation}
where
$\mathbf{B}\!\in\!\mathbb{R}^{d\times r}$ and
$\mathbf{A}\!\in\!\mathbb{R}^{r\times k}$,
and the rank satisfies $r \ll \min(d,k)$.
During training, \emph{only} $\mathbf{A}$ and $\mathbf{B}$ are updated,
while $\mathbf{W}_{0}$ remains fixed.
\\
\textbf{Forward pass.}\;
Given a hidden vector $\mathbf{x}$,
a LoRA-augmented linear layer computes
\begin{equation}
    \mathbf{h}
    \;=\;
    \mathbf{W}_{0}\,\mathbf{x}
    \;+\;
    \mathbf{B}\bigl(\mathbf{A}\mathbf{x}\bigr),
    \label{eq:lora_forward}
\end{equation}
adding at most $r(d{+}k)$ trainable parameters—
typically \mbox{$\!{<}2\%$} of the layer’s original size.

\vspace{0.2em}
\noindent
\textbf{Why LoRA suits continual learning.}\;
\begin{enumerate}[label=(\alph*)]
  \item \textit{Tiny footprint.}   Typically using small rank values achieves the task accuracy required on par with the full fine-tuning of the model ($r\!=\!1\text{--}4$ in our experiments). This helps in keeping the memory, compute overhead negligible. Specifically, $\!{<}2\%$ additional parameters per task, to train dozens of tasks on a single GPU.
  \item \textit{Task Modularity.}  An independent LoRA adapter is allocated per task, preventing interference without touching the backbone. This leads to acquisition of new knowledge without interference with existing knowledge. Recent work shows task-specific LoRA weights can be merged, hot-swapped, re-used
at inference time for different usecases.
  \item \textit{Compatibility with compression.}  LoRA architecture perfectly suits the proposed approach of injecting the adapter into the latent space of the auto encoder. After training on a task we encode the pair $\{\mathbf{A},\mathbf{B}\}$ with our conditional auto-encoder (\ref{Sec:AE}), store only the latent code, and discard the full low-rank weights. Thus LoRA serves as a lightweight \emph{front-end} for task adaptation, while CAE provides long-term storage.
\end{enumerate}

\subsection{Auto Encoder}
\label{Sec:AE}
Our framework is not tied to any specific autoencoder design.  
In practice, we employ \textbf{contractive autoencoders (CAEs)} \cite{salah2011contractive} because
they consistently proved more stable than alternatives such as variational
autoencoders.  A CAE replicates the architecture of a standard autoencoder \cite{doersch2016tutorial}
but augments the loss with a contraction term that discourages large
perturbations in the latent space:

\begin{equation}
  \mathcal{L}_{\text{CAE}}(\theta) \;=\;
  \cos\!\bigl(\theta,\hat{\theta}\bigr)
  + \lambda \,\lVert J_{f}(\theta) \rVert_F^{2},
  \label{eq:cae_loss}
\end{equation}

where the first term is the cosine-similarity reconstruction loss
and the second term is a contractive
penalty.  The penalty is the Frobenius norm of the Jacobian of the
encoder’s hidden representation with respect to each input~$x_i$:

\begin{equation}
  \lVert J_{f}(x) \rVert_F^{2}
  \;=\;
  \sum_{i,j}
  \left(
    \frac{\partial h_j(\theta)}{\partial x_i}
  \right)^{2},
  \label{eq:contractive_penalty}
\end{equation}

with $h_j(\theta)$ denoting the activation of the $j$-th hidden unit.
Throughout our experiments we set the regularization coefficient to
$\lambda = 1\times10^{-4}$.

\subsection{Framework Overview}

Figure~\ref{fig:framework} sketches our \textbf{COLA framework} for
lifelong learning without catastrophic forgetting. As shown in the fig each new task is handled in three stages—\emph{adapt}, \emph{compress}, and \emph{replay on demand}(during inference)—all while keeping the backbone frozen and storing \emph{no raw data} from
previous tasks.

\vspace{0.4em}
\noindent\textbf{Stage 1: Task adaptation.}\; When a new task~$T_i$ arrives where each task is specified as ${\mathcal D}_i$ $ {\mathcal D}_i=\{(X^{(j)},y^{(j)})\}_{j=1}^{n_i}$ which consists of n different training pairs. Our framework incorporate a  rank-$r$ LoRA matrices adapter into  pre-trained frozen backbone model. We fine-tune the frozen backbone fitted with its freshly attached adapter on the current task’s dataset ${\mathcal D}_i$ for task adaptaion. 
After adaptation, the learned low-rank matrices $(\mathbf{A}_i,\mathbf{B}_i)$ are vectorized into a compact \emph{adapter snapshot}
$\boldsymbol{\theta}_i=\operatorname{vec}[\mathbf{A}_i;\mathbf{B}_i]$. 


\noindent\textbf{Stage 2: Lifelong encoding.}\;
As shown in Fig \ref{fig:framework} for the next stage we employ a CAE for lifelong encoding. We train a \emph{Lifelong Auto-Encoder} (LAE)
to encode each adapter into a fixed-size latent vector  $\mathbf{z}_i\in\mathbb{R}^m$ as shown in the Eq \ref{eq:autoencoder} where $f$ indicates the encoder, $g$ indicates the decoder of the autoencoder and overall autoencoder is optiomized using the contractive loss as shown in Eq \ref{eq:cae_loss}.

\begin{equation}
  \mathbf{z}_i = f_{\phi}(\boldsymbol{\theta}_i),
  \quad
  \hat{\boldsymbol{\theta}}_i = g_{\psi}(\mathbf{z}_i),
  \label{eq:autoencoder}
\end{equation}


The CAE therefore learns a contractive latent manifold that
captures the intrinsic geometry of LoRA updates while keeping the code
dimension $m\!\ll\!\lvert\boldsymbol{\theta}_i\rvert$.
After convergence we discard discard the \emph{encoder} $f_\phi$ and store only the decoder $g_\psi$ plus the latent vectors $\{\mathbf{z}_i\}$.  This allows us to reconstruct any adapter on demand (via $g_\psi(\mathbf{z}_i)$) while reducing persistent storage by over an order of magnitude. After each task adaptation, its adapter snapshot $\boldsymbol{\theta}_i$ may be immediately sent to the LAE for encoding or temporarily held in a small FIFO buffer of size $M$.  Whenever the buffer fills—or at scheduled intervals—we batch‐encode all stored adapters via the CAE and then purge the raw weight tensors, ensuring stable auto‐encoder training while smoothing out per‐task encoding costs.


\vspace{0.4em}

\noindent\textbf{Stage 3: Adapter selection at inference.}\;
In a \emph{hard} continual-learning setup, the task identity is \emph{not} given at test time.  At inference, we therefore proceed as follows:

\begin{enumerate}[label=(\roman*)]
  \item If a task-ID $i$ is \emph{provided}, we simply retrieve its latent code $\mathbf{z}_i$, decode it via $g_{\psi}$ to reconstruct adapter $\hat{\boldsymbol{\theta}}_i$, install that adapter in the frozen backbone, and generate outputs directly.
  \item Otherwise (no task-ID available), we must automatically choose among all $K$ stored adapters:
    \begin{enumerate}[label=(\alph*)]
      \item Decode each latent vector $\mathbf{z}_t$ into adapter weights $\hat{\boldsymbol{\theta}}_t = g_{\psi}(\mathbf{z}_t)$.
      \item For each adapter $t$, compute the model’s \emph{perplexity} on the input sequence $X$:
      \[
        \mathrm{PPL}_t(X)
        \;=\;
        \exp\Bigl(-\tfrac{1}{|X|}\sum_{i=1}^{|X|}\log p_{\theta_t}(x_i\mid x_{<i})\Bigr),
      \]
      where $p_{\theta_t}$ are the token probabilities under adapter $t$.
      \item Select the adapter index
      \[
        \hat t \;=\;\arg\min_{1\le t\le K}\mathrm{PPL}_t(X),
      \]
      install $\hat{\boldsymbol{\theta}}_{\hat t}$, and produce the response.
    \end{enumerate}
\end{enumerate}

This scheme trades off $K$ forward passes for a built-in confidence estimate, avoiding the need for any extra classification head (and its associated forgetting risk) while still operating under realistic “no task-label” constraints.The CAE decoder and LM forward passes for different adapters are independent, we can \emph{batch} reconstruct all $K$ adapters and score them in parallel, making the total added latency negligibl while only ever instantiating one adapter’s weights at a time for generation.


\vspace{0.4em}

\noindent\textbf{Optional Warm-start via adapter reuse for transfer learning}\;
When a new task arrives, we first compute its perplexity under each existing adapter (as in Stage 3) and select the best‐scoring adapter $\hat t$.  We then initialize the new task’s LoRA matrices $(\mathbf{A}_i,\mathbf{B}_i)$ from $\hat{\boldsymbol{\theta}}_{\hat t}$ rather than from random, effectively using the most relevant adapter as a \emph{pre‐trained warm start}.  This transfer‐learning step accelerates convergence and improves final accuracy, as evidenced by the speed‐up curves in Fig \ref{fig:forwardtransfer}.

\vspace{0.4em}
\noindent\textbf{Key properties.}\;
(i)~\emph{Constant storage}: the memory footprint grows with the number
of \emph{latents} not with full adapters.
(ii)~\emph{No rehearsal}: the LAE operates on weights, not examples; no
dialogue data are cached.
(iii)~\emph{Task modularity}: the frozen backbone and per-task codes
cleanly separate generic linguistic knowledge from domain-specific
behaviour, enabling plug-and-play continual learning without
catastrophic forgetting.
(iv)~\emph{Warm-start transfer:} new tasks can initialize from the best existing adapter (via perplexity-based selection), speeding convergence and boosting performance.  

%% file: Sections/3_ExpANDresults.tex
\section{Experiments} To validate our approach, we performed experiments on we carried out a wide range of Continual Learning experiments on a variety of datasets to to test for  class incremental learning and Intent detection in Task Orient Dialogue systems, we empirically established the level of precision needed when approximating of a task network in order to achieve comparable performance on a task. Secondly, we also verified that our proposed framework can reconstruct large of task networks to test the system’s ability to encode a very large number of tasks, thus validating that the proposed approach does not merely memorize task-specific networks. Finally, we assessed our method's performance on the following benchmarks.

\subsection{Datasets}
We experiment our method on different scenarios such as Intent Classification, dialogue state tracking (DST), Natural Language Generation(NLG) and end-to-end response generation (E2E). We performed experiments on four different 
datasets— Task-Master 2019 (TM19) \cite{byrne2019taskmaster},
Task-Master 2020 (TM20) \cite{byrne2019taskmaster},
Schema Guided Dialogue (SGD) \cite{rastogi2020towards} and MultiWoZ \cite{zang2020multiwoz,eric2019multiwoz} following the methodology defined in Section \ref{sec:methodology}. Below, we first present our dataset and implementation details, then discuss our results.

We experiment our methods with the following datasets. \\

\textbf{CLINC150} This dataset \cite{larson2019evaluation} is a multi-domain dataset
that contains 23,700 utterances used for evaluation of intent classification and out of scope prediction over wide range of intents. Overall this dataset cover 150 intent classes over 10 domains.\\
\textbf{MultiWOZ} This dataset \cite{budzianowski2018multiwoz} craved using Wizard-of-Oz (WOZ) setup. MultiWOZ compraises of multi-turn dialogues,which is widely used to evaluate the task oriented dialogue(TOD) systems. Datasets contains of domains like restaurant,..etc. There are two versions of this dataset: Initial version proposed by \cite{zang2020multiwoz} and later  \cite{eric2019multiwoz} corrected version the addressing the  errors and inconsistencies. We also use task oriented datasets such as Task-Master 2019 (TM19) \cite{byrne2019taskmaster},
Task-Master 2020 (TM20) \cite{byrne2019taskmaster},
Schema Guided Dialogue (SGD) \cite{rastogi2020towards}\\
\textbf{Task Master}: This datasets consists of corpus of 55,000 spoken and written task-oriented dialogues in over a different domains. Task-Master 2019 (TM19) \cite{byrne2019taskmaster} contains 6 domains eg: restaurants, food ordering, moviews ..etc,
Task-Master 2020 (TM20) \cite{byrne2019taskmaster} contains 23,789 movie ticketing dialogues which span over domains such as deciding on theater, time, movie name, number of tickets, and date, or opt out of the transaction..etc.\\
\textbf{Schema-Guided Dialogue (SGD)} This dataset\cite{rastogi2020towards} contains over 20k annotated examples spanning over 20 domains such as s banks, events, media, calendar..etc. These annotated examples are the interactions between the Virtual assistant and human over services and API's.\\

We also utilize a combines dataset Natural Language understanding that is a relabeled and aggregated version of three
large NLU corpuses: CLINC150 \cite{larson2019evaluation} ,
Task-Master 2020 (TM20) \cite{byrne2019taskmaster},
Schema Guided Dialogue (SGD) \cite{rastogi2020towards} and MultiWOZ \cite{budzianowski2018multiwoz}.



\begin{table}[h!]
\centering
\begin{tabular}{lrrrrrr}
\hline
\textbf{Name} & \textbf{Train} & \textbf{Valid} & \textbf{Test} & \textbf{Dom.} & \textbf{Intent} & \textbf{Turns} \\ \hline
TM19          & 4,403          & 551            & 553           & 6             & 112             & 19.97          \\
TM20          & 13,839         & 1,731          & 1,734         & 7             & 128             & 16.92          \\
MWoZ          & 7,906          & 1,000          & 1,000         & 5             & 15              & 13.93          \\
SGD           & 5,278          & 761            & 1,531         & 19            & 43              & 14.71          \\

CLINIC150           & 10,525         & 3,080            & 5,470         & -            & 10              & -     \\ \hline

\textbf{Total} & 41,951         & 7,123       & 10,288         & 37            & 290             & 16.23          \\ \hline

\end{tabular}
\label{tab:dataset_statistics}
\caption{Main datasets statistics.}
\end{table}
\subsection{Implementaion Details}

\subsubsection{State of Art Comparison and Metrics}
For experiments on Task-Master 2019, Task-Master 2020 (TM20) \cite{byrne2019taskmaster},
Schema Guided Dialogue (SGD) \cite{rastogi2020towards} and MultiWOZ \cite{budzianowski2018multiwoz}  we used a decoder-only model GPT-2 where froze the model parameters and updated only the adapter parameters. For adapter, we used bottleneck sizes of bottleneck size b between 10, 50, 100, 200. We used an auto encoder with latent vector of size 50. 

\textbf{Metrics:} For \textit{Intent Recognition}: We measure the precision of intent prediction by comparing the predicted intent of the model with the gold standard label. For \textit{Dialogue State Tracking (DST)} the performance is quantified using Joint Goal Accuracy (JGA) as define as the percentage of dialogue turns where all slot-value pairs match the gold state. and for \textit{Natural Language Generation (NLG)}: Quality is assessed with the BLEU metric \cite{papineni2002bleu} along with the slot error rate (SER \cite{madotto2020continual}), calculated as the fraction of required slot values that are missing or incorrect in the generated response.

\textbf{Baselines:} We compare our method across various existing  
approaches.
\begin{itemize}
    
    \item \textbf{Adapter-style \cite{madotto2020continual}}: inserts lightweight adapter modules into the base model; each task trains its own adapter, enabling modular task-specific learning without overwriting shared model weights.

    \item \textbf{Replay (GEM) \cite{chaudhry2018efficient}}: fine-tunes the entire model while storing and replaying real samples from prior tasks via a memory buffer; uses gradient projection to avoid catastrophic forgetting.
    
    \item \textbf{EWC \cite{kirkpatrick2017overcoming}}: fine-tunes the model with an additional regularization term that penalizes changes to parameters important for previous tasks, based on the Fisher Information Matrix.
    
    \item \textbf{L2 Regularization \cite{hoerl1970ridge}}: constrains the model by penalizing deviation from previously learned weights using a simple L2 loss, encouraging retention of earlier knowledge.
    
    \item \textbf{Vanilla FT}: sequentially fine-tunes the entire model on each task without any regularization or memory, typically resulting in catastrophic forgetting.
\end{itemize}

\subsubsection{Results} Table \ref{tab:e2e_results} presents a performance comparison of our method against various continual learning methods across multiple CL datasets. First, we trained a LoRA adapter attached to pre-trained gpt2 network, in which only LoRA parameters are updated for learning the task while rest of the network remains constant. The LoRA network is flatted and splitted to fed into our Our AE which consists of two connected layers with 100,000, 50 parameters respectively.  Thus, our latent vectors were of size 50. Each task network can be ingested into auto encoder in a sequential fashion.

We report results from average of five random runs with different task orders for each run on the CL benchmark. In fairness, we used the same rank of the LoRA rank adapter in each corresponding comparison experiment that utilize adapter-style learning. Across all tasks on different datasets of the standard CL benchmark, Our method consistently outperforms previous methods by a consistent margin and  on par with task-specific models trained independently, and highlighting the effectiveness of our approach. Our methods also significantly surpass several state-of-the-art continual learning approaches, including EWC, L2, and AGEM, LAMOL and replay based methods. Unlike these baselines, our methods retain knowledge of previous tasks while maintaining the ability to learn new ones effectively.

\begin{figure}[t]
  \centering
  \includegraphics[width=0.8\linewidth]{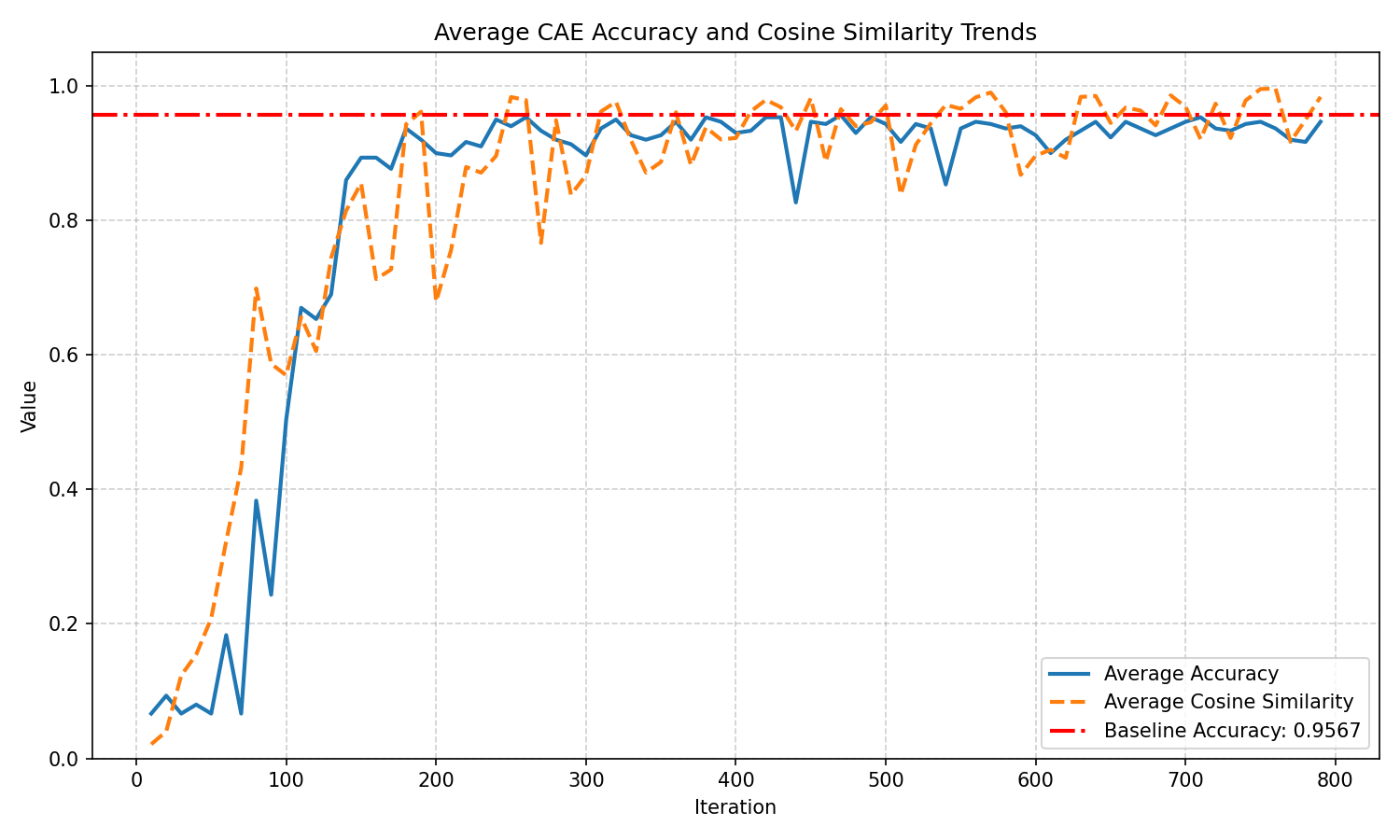}
  \caption{COLA: Avg Accuracy vs Baseline vs Reconstruction Score}
  \label{fig:cosinevsaccuracy}
\end{figure}

We evaluated the average task performance relative to changes in model parameters generated by the auto encoder, quantified using reconstruction score.  As shown in Fig \ref{fig:cosinevsaccuracy} results reveal a strong correlation: as reconstruction similarity score decreases, performance tends to decrease. The baseline accuracy is shown as a dotted red line,  marks the point where the  which model performance closely matches that of the original. Therefore, we use a threshold of 0.99 as the stopping criterion in our following experiments, unless otherwise specified.

\begin{table}
\centering
\begin{tabular}{lcccccc}
\hline
\textbf{Method} & \textbf{Accuracy↑} & \textbf{JGA↑} & \textbf{EER↓} & \textbf{BLEU↑} \\
\hline
\textit{VANILLA} &  5.12  & 8.91 & 50.76  & 7.48  \\
\textit{L2 \cite{hoerl1970ridge}} & 4.81  & 7.21 & 60.88  & 5.8 \\
\textit{EWC \cite{kirkpatrick2017overcoming}} & 4.85  & 9.56 & 61.33  & 5.56  \\
\textit{AGEM \cite{chaudhry2018efficient}} & 39.14  & 12.73 & 67.99  & 4.64  \\
\textit{LAMOL \cite{sun2019lamol}} &  12.59  & 8.45 & 71.12  & 3.2  \\
\textit{REPLAY \cite{robins1995catastrophic}} &  82.38  & 30.32  & \textbf{16.12 } & \textbf{17.9} \\
\textit{Ours} &  \textbf{89.87} & \textbf{34.86} & 32.58  & 16.26  \\
\hline
\textit{MULTI} &  93.45 & 47.3  & 11.46  & 22.51  \\
\hline
\end{tabular}
\label{tab:e2e_results}
\caption{E2E results in terms of Intent accuracy, Joint Goal Accuracy (JGA), Slot Error Rate (EER), and BLEU. \textbf{+Param} indicates additional parameters per task, and \textbf{Mem} indicates episodic memory per task.}
\end{table}

\begin{figure}[t]
  \centering
   \includegraphics[width=1.0\linewidth]{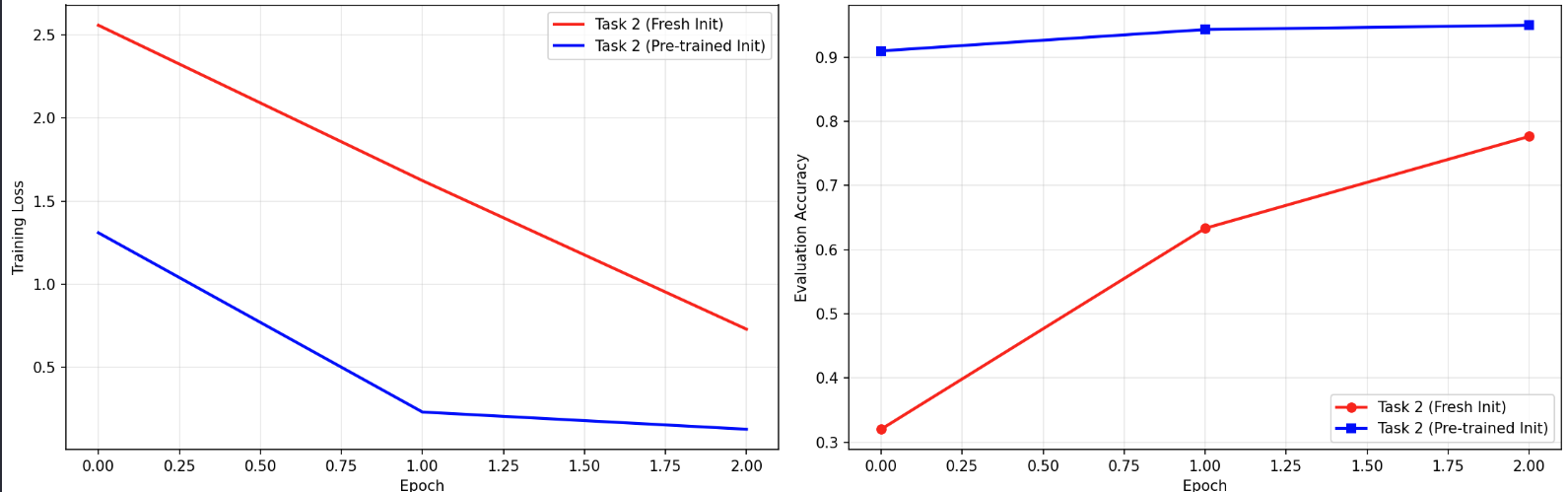}
    \caption{COLA: Avg Accuracy vs Baseline vs Reconstruction Score}
   \label{fig:forwardtransfer}
\end{figure}





\subsection{Continual Learning Transfer Validation}

To validate the effectiveness of our continual learning approach, we conducted a controlled experiment examining knowledge transfer between sequential tasks using the CLINC-OOS intent classification dataset. The experiment consisted of three phases: (1) training a RoBERTa-base model with LoRA adapters on Task 1 (achieving 95.33\% accuracy), (2) training a fresh model on Task 2 from scratch, and (3) initializing Task 2 training with the pre-trained Task 1 model weights. Our results demonstrate significant evidence for positive knowledge transfer in continual learning scenarios. The model pre-trained on Task 1 achieved 95.00\% accuracy on Task 2, representing a substantial 17.33 percentage point improvement over the fresh model trained from scratch (77.67\% accuracy). This improvement indicates that the learned representations from Task 1 contain transferable knowledge that accelerates learning and enhances performance on related downstream tasks.

The experimental validation confirms that continual learning through parameter-efficient fine-tuning enables effective knowledge reuse across sequential tasks. The pre-trained model not only achieved superior final performance but also demonstrated faster convergence, reaching high accuracy levels within the same training epochs as the baseline. This finding supports the fundamental hypothesis that neural networks can accumulate and leverage knowledge from previous tasks to improve learning efficiency on new but related problems. The success of this transfer learning approach, particularly in the context of intent classification where tasks share underlying linguistic patterns and semantic structures, provides empirical evidence for the viability of continual learning systems in practical NLP applications.

\subsection{Parameter and Memory Overhead in Continual Learning Approaches}

Table~\ref{tab:cl_storage} presents a comparative analysis of various continual learning (CL) strategies in terms of their storage overhead during both training and deployment. The comparison includes replay-based, regularization-based, and modular adaptation approaches.

Our method, CAE Replay, stands out as a highly efficient solution. During training, it requires storing only the decoder parameters ($n\phi$) in addition to the shared base model ($\Theta$), avoiding the need to retain large memory buffers or multiple full model copies. At deployment time, the advantage becomes even more pronounced: unlike replay-based methods such as Replay or GEM, which must retain the full memory buffer ($n|M|$), and adapter-based methods that store task-specific modules ($n|\mu|$), CAE Replay only retains compact task-specific decoders.

This significantly reduces storage cost without compromising task-specific adaptability. The method is especially suitable for deployment scenarios with limited memory capacity, such as edge devices or large-scale distributed systems, where lightweight task modules are critical. Furthermore, unlike regularization-based methods (e.g., EWC or L2), which are storage-efficient but prone to catastrophic forgetting, CAE Replay retains strong task performance by leveraging its generative memory mechanism through compact decoders.

\begin{table}
\centering
\begin{tabular}{lccc}
\hline
\textbf{Method}     & \textbf{Train Overhead} & \textbf{Deploy Overhead} & \textbf{Notes}           \\ \hline
CAE Replay          & $\Theta + n\phi$        & $n\phi$                  & \cmark\ Lightweight      \\
Replay / GEM        & $\Theta + n|M|$         & $\Theta + n|M|$          & \xmark\ High memory      \\
Adapter-style       & $\Theta + n|\mu|$       & $n|\mu|$                 & \cmark\ Modular          \\
EWC                 & $\Theta$                & $\Theta$                 & \warnmark\ Forgetting?   \\
L2 Reg.             & $\Theta$                & $\Theta$                 & \warnmark\ Forgetting?   \\
Vanilla             & $\Theta$                & $\Theta$                 & \xmark\ No memory        \\ \hline
\end{tabular}
\label{tab:cl_storage}
\caption{Comparison of storage overhead across continual learning methods.}
\end{table}


%% file: Sections/2b_Related_work.tex
\section{Related Work}
Continual Learning a.k.a Lifelong Learning aims to solve the problem of achieving good  performance on alot of numbers of tasks that are arriving sequentially without catastrophic forgetting \cite{kirkpatrick2017overcoming,mccloskey1989catastrophic}. Based on the nature of the tasks, Continual learning can be broadly divided into three categories: Class-Incremental Learning (CIL), Domain-Incremental Learning (DIL), Task Incremental Learning (TIL). 
\begin{itemize}
  \item \textbf{Class-Incremental Learning (CIL)}  
    The model must continually expand its set of recognizable classes and correctly classify inputs among all seen classes.
  \item \textbf{Domain-Incremental Learning (DIL)}  
    Methods that fine-tune large language models on a stream of domain-specific instructions, improving their ability to perform tasks within a particular subject area.
  \item \textbf{Task-Incremental Learning (TIL)}  
    Approaches that sequentially fine-tune a model on each new task—knowing task boundaries—so that it acquires the capability to solve each distinct task as it arrives.
\end{itemize}

Depending on how each task’s knowledge is stored and subsequently leveraged as new tasks arrive, task-incremental methods can be divided into three categories. 1) \textit{Regularization-based} : where regularization based constraints are introduced on model parameters by modifying the loss function to overcome catastrophic forgetting while learning new tasks. Some of the methods include EWC ..etc \cite{chaudhry2018riemannian,aljundi2018memory,kirkpatrick2017overcoming} 2. \textit{Reply-Based methods} where data from previous tasks either in the form of exemplars or buffer or complete dataset are explicitly stored and revisited to learn new things. Key appraoches include Experience Replay\cite{playing_atari_deep_reinforcement}, iCarl \cite{iCarl} and some othe methods include \cite{arani2022learning,caccia2021new,variational_continual_learning,bonicelli2022effectiveness,sarfraz2023error,Gradient_episodic_memory,wang2023dualhsic,liang2023loss}.  3.\textit{Architecture-based Networks} where model architecture is evolute as the model training goes on while learning new task. This happens either expanding the network to include new knowledge \cite{mallya2018packnet,progressive_networks,dynamically_expandable_networks,context_dependent_gating,serra2018overcoming,hung2019compacting,jayamandivarapu,camp2020continuallearningdeepartificial}

Several recent studies have shown that fine-tuning only a small portion of a model’s parameters can match the performance of full-model updates \cite{karimi2021compacter,lester2021power,li2021prefix}. Although most of this work targets single-task learning, a few efforts have begun to apply parameter-efficient approaches within continual learning. For example,\cite{madotto2020continual} introduce AdapterCL, which assigns a unique adapter module to each new task.However, these approaches either demand the continual growth of adapter modules or maintain a fixed adapter architecture, which in turn limits their capacity to integrate new information.

%% file: Sections/5_conclusion.tex
\section{Conclusion}
In this work we reframed continual learning, and more generally task-oriented dialogue learning,
as a \textit{generative} problem solvable by a frozen language-model backbone equipped with LoRA adapters.
Our methodology couples a pre-trained base model with \textsc{MixAdapter} blocks, allowing the system to acquire new skills sequentially while leaving the backbone untouched.
Beyond simplifying multi-task training, adapter isolation highlights the classic trade-off
among parameter count, replay memory, and model capacity: storing knowledge in many small bottlenecks
removes the need for large episodic buffers yet still preserves prior competence.
This finding underscores a broader \textit{no-free-lunch} principle in lifelong dialogue systems:
memory efficiency, model plasticity, and final performance
must be balanced rather than optimized in isolation. Our experiments demonstrate that the proposed framework
\emph{efficiently acquires and retains a large number of tasks}
without catastrophic forgetting, validating its effectiveness
for real-world continual dialogue scenarios.